# Viability of Low-Cost Infrared Sensors for Short Range Tracking

**Noah Haeske**

ABSTRACT

A classic task in robotics is tracking a target in the external environment. There are several well-documented approaches to this problem. This paper presents a novel approach to this problem using infrared time of flight sensors. The use of infrared time of flight sensors is not common as a tracking approach, typically used for simple motion detectors. However, with the approach highlighted in this paper they can be used to accurately track the position of a moving subject. Traditional approaches to the tracking problem often include cameras, or ultrasonic sensors. These approaches can be expensive and overcompensating in some use cases. The method focused on in this paper can be superior in terms of cost and simplicity.

## Introduction

The sensor highlighted in this paper is the **VL53L7CX** by ST. This sensor was selected for its wide field of view (see methods) but is comparable to most other Infrared TOF sensors. This sensor works by emitting a 940nm wavelength light in the direction it is facing. The on-board microcontroller then begins a clock. The light is reflected off the target as well as other surroundings, and a fraction of it is returned to the sensor. The sensor contains an array of sixty-four single photon avalanche diodes (SPADs). Each of these diodes, when triggered by the reflected infrared light, stops the clock, and returns a time value. This time value can then be multiplied by the speed of light in order to calculate the distance the light traveled. This distance is then divided by two in order to find the distance from the sensor to the target.

$$\frac{1}{2}(t \times c) = d$$

## Sensor Comparison

While not a traditional method of detection, in certain short-range applications infrared TOF sensors are an appealing option. In comparison to cameras and computer vision, infrared TOF sensors are significantly cheaper and require less space. In comparison to ultrasonic sensors, infrared TOF sensors offer a wider field of view (Adarsh et al, 2016, p. 2). This is significant for small builds, where four or more ultrasonic sensors could be replaced by two infrared sensors.

## Ambient Lighting

A known potential drawback of the approach used in this paper is the effects of ambient lighting. As previously mentioned, the ultrasonic sensor relies on an array of sixty-four SPADs to measure distance. These diodes are specifically designed to pick up on the 940nm wavelength infrared light emitted by the sensor. However in conditions with strong light coming from the external environment, the SPADs can detect light that did not originate from the sensor. This is a common source of error for infrared sensors.

# Triangulation

As previously discussed, infrared TOF sensors return an array of distance values. However, these distance values on their own are not enough to derive position. While there are multiple ways to combine sensors in order to derive position, the method highlighted in this paper is a form of triangulation. In order to calculate position, two sensors placed in different positions along the same axis must scan the same area. The lowest distance value returned by each sensor is then stored in a program. The known distance between the two sensors is also stored in the program. These three distance values allow for the mathematical formulation of a triangle, with the distance between the two sensors as the base. The following equations are then used to find the (x, y) coordinates of the target.

$$\theta = \cos^{-1}\frac{(A^2 + C^2 - B^2)}{(2 \times A \times C)}$$

$$x_2 = x_1 + (A \times \cos\theta)$$

$$y_2 = y_1 + (A \times \sin\theta)$$

In this these equations:

A: The lowest returned distance from sensor one
B: The lowest returned distance from sensor two
C: The known distance from sensor one to sensor two
$\theta$: The angle between the axis the sensors are on and the target
$x_1$: The x coordinate of sensor one
$y_1$: The y coordinate of sensor one
$x_2$: The x coordinate of the target
$y_2$: The y coordinate of the target

# Methods

In order to perform repeatable tests using this sensor, a test frame was constructed. To simulate the close range tracking this paper is focused on, the test frame was built in a 1m x 1m square, with one side missing to allow a subject to move in and out of the frame. Furthermore, to ensure the sensors remained in the same position through all the tests, a specialized holding bracket was modelled, and 3D printed. (see Fig. 1) The bracket allows the sensor to sit at a constant angle. It also incorporates holes for wiring and screws. Two of these brackets were installed in adjacent corners of the frame. (see Fig. 2)

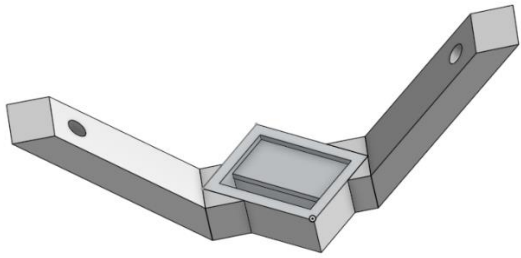
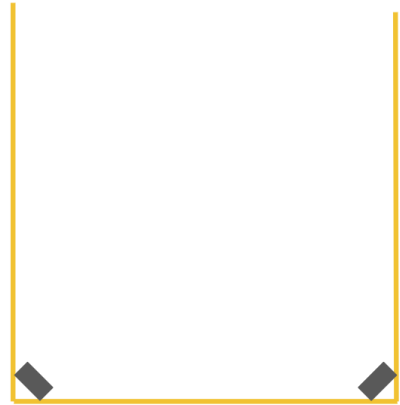

Figure 1        Figure 2

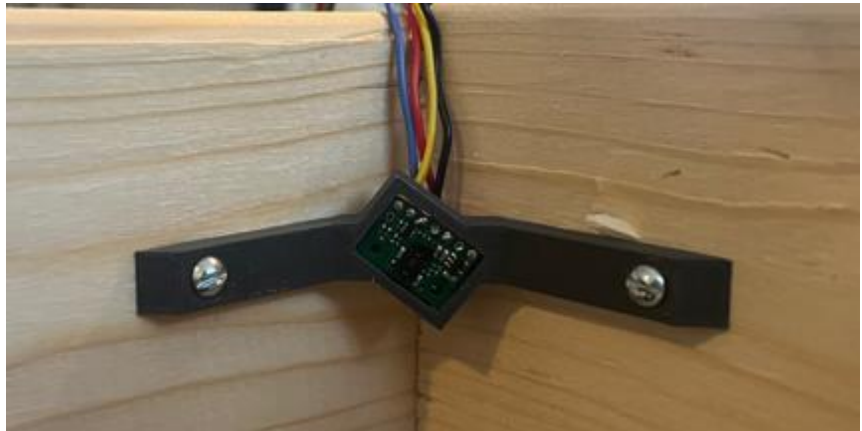

Figure 3

**Figure 1.** CAD model of sensor mounting bracket.

**Figure 2.** Drawing of test frame. The outer lines represent 1-meter lengths of wood. Gray regions represent mounting brackets.

**Figure 3.** The mounted sensor in the bracket.

A program was written in order to drive the sensors and interpret data. The script uses python as well as ST's library for the sensor. The program was run on a raspberry pi, to which the sensors were connected via I2C connections. A graphical interface was also created in order to debug the program. The graphical interface included a four by four heatmap from each sensor, as well as a marker indicating estimated position of the target. Code is available at https://github.com/noah-haeske/research

In order to determine accuracy of the system, two experiments were conducted. In each experiment four locations were measured out and marked within the frame. A measuring target was created from a tripod attached to a 10 cm diameter foam cylinder. The stand was covered in fabric to simulate a clothed human limb. The actual measured position of the stand was recorded. The system was then run for 10 trials (~10 seconds) and the reported position of the stand was recorded. The stand was then moved to the next marked location and all steps were repeated. This experiment was performed under two different ambient lighting conditions: no ambient light, and artificial lighting.

# Results

After conducting the previously mentioned experiments, the results were as follows.

## Experiment One

Experiment one was performed in artificial ambient lighting. Ten trials were performed in each of the target's four set positions. The recorded measurements can be found below. All values are in millimeters.

**Table 1.**

| Trial | Position 1 x | Position 1 y | Position 2 x | Position 2 y | Position 3 x | Position 3 y | Position 4 x | Position 4 y |
|---|---|---|---|---|---|---|---|---|
| 1 | 304 | 298 | 647 | 267 | 659 | 715 | 300 | 751 |
| 2 | 329 | 266 | 640 | 263 | 654 | 769 | 242 | 767 |
| 3 | 305 | 278 | 636 | 261 | 804 | 666 | 204 | 787 |
| 4 | 307 | 269 | 637 | 266 | 640 | 734 | 152 | 781 |
| 5 | 295 | 297 | 637 | 270 | 636 | 728 | 203 | 785 |
| 6 | 302 | 296 | 634 | 247 | 661 | 745 | 226 | 774 |
| 7 | 317 | 258 | 643 | 273 | 652 | 763 | 247 | 768 |
| 8 | 319 | 289 | 625 | 238 | 630 | 664 | 225 | 770 |
| 9 | 315 | 276 | 628 | 243 | 645 | 667 | 250 | 772 |
| 10 | 292 | 294 | 637 | 253 | 526 | 725 | 210 | 770 |
| Actual | 330 | 330 | 660 | 330 | 330 | 660 | 660 | 660 |

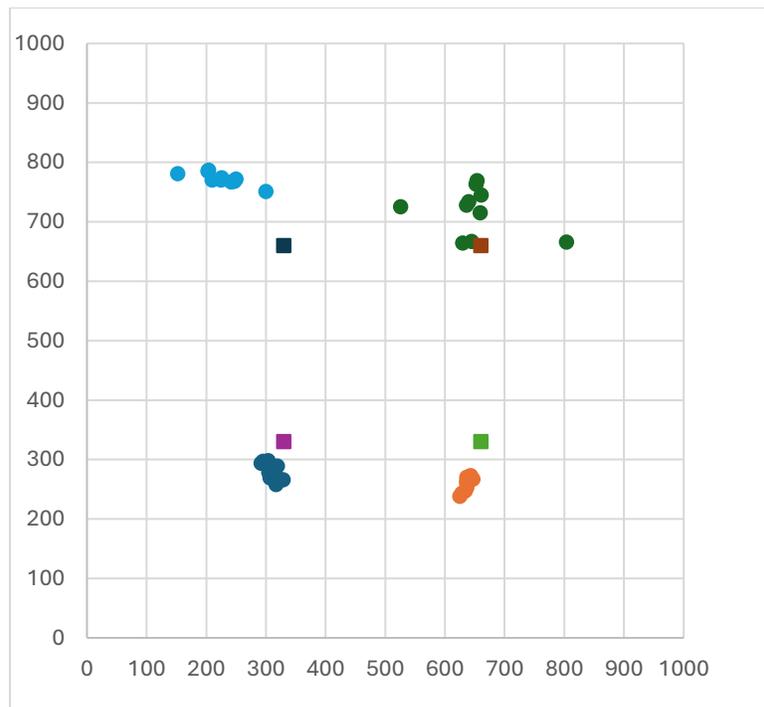

Figure 3

**Figure 3.** A scatterplot of the measured values from experiment 1. Each color represents a trial with the target placed in a different position. Each dot represents one trial of the program. Each square represents an actual position of the target. All measurements are in millimeters.

## Experiment Two

Experiment two was performed with no ambient lighting. Ten trials were performed in each of the targets four set positions. The recorded measurements can be found below. All values are in millimeters

Table 2.

| Trial | Position 1 x | Position 1 y | Position 2 x | Position 2 y | Position 3 x | Position 3 y | Position 4 x | Position 4 y |
|---|---|---|---|---|---|---|---|---|
| 1 | 314 | 324 | 648 | 294 | 645 | 675 | 300 | 669 |
| 2 | 315 | 304 | 638 | 296 | 686 | 680 | 342 | 661 |
| 3 | 315 | 295 | 646 | 295 | 662 | 666 | 335 | 680 |
| 4 | 317 | 289 | 636 | 292 | 670 | 673 | 311 | 682 |
| 5 | 299 | 332 | 647 | 301 | 647 | 682 | 305 | 673 |
| 6 | 317 | 299 | 636 | 296 | 655 | 690 | 347 | 649 |
| 7 | 327 | 288 | 640 | 303 | 652 | 687 | 343 | 648 |
| 8 | 335 | 326 | 635 | 298 | 657 | 683 | 347 | 671 |
| 9 | 317 | 306 | 631 | 293 | 671 | 684 | 323 | 682 |
| 10 | 312 | 309 | 647 | 305 | 649 | 675 | 301 | 689 |
| Actual | 330 | 330 | 660 | 330 | 660 | 660 | 330 | 660 |

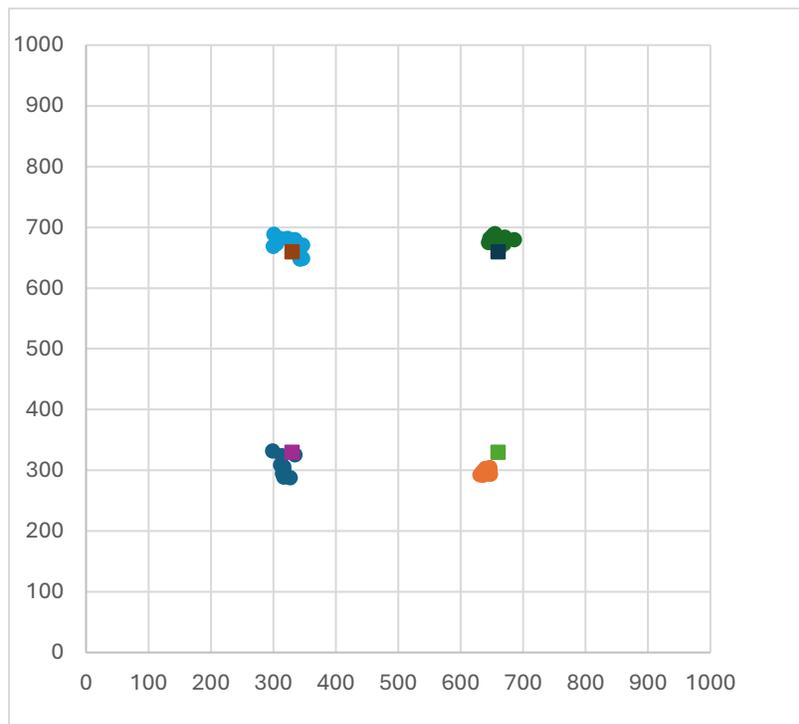

Figure 4

**Figure 4.** A scatterplot of the measured values from experiment two. Each color represents a trial with the target placed in a different position. Each dot represents one trial of the program. Each square represents an actual position of the target. All measurements are in millimeters.

## Further Calculations

Further calculations were done in order to enhance understanding of the systems precision and accuracy. The standard deviation between each of the ten trials performed in each position were calculated. The standard deviation is based on the grouping of the trials, and can be used to track precision across the experiments. The data is shown below.

**Table 3.**

| Standard Deviation | Position 1 x | Position 1 y | Position 2 x | Position 2 y | Position 3 x | Position 3 y | Position 4 x | Position 4 y | Average |
|---|---|---|---|---|---|---|---|---|---|
| Exp. 1 | 10.85 | 13.87 | 6.14 | 11.48 | 63.27 | 37.40 | 36.74 | 9.83 | 23.70 |
| Exp. 2 | 8.89 | 14.77 | 5.82 | 4.15 | 12.29 | 6.86 | 18.66 | 13.28 | 10.59 |

Furthermore, the percentage error for each position in each experiment was calculated. The percentage error is based on the actual position of the target and can be used to track accuracy across the experiments. The data is shown below

**Table 4.**

| Percent Error | Position 1 x | Position 1 y | Position 2 x | Position 2 y | Position 3 x | Position 3 y | Position 4 x | Position 4 y | Average |
|---|---|---|---|---|---|---|---|---|---|
| Exp. 1 | 6.52% | 14.52% | 3.58% | 21.79% | 97.18% | 8.73% | 65.77% | 17.05% | 29.39% |
| Exp. 2 | 4.30% | 7.03% | 2.97% | 9.91% | 1.58% | 2.95% | 5.27% | 2.27% | 4.54% |

## Discussion

The data from experiments one and two lead to interesting discoveries about the system. Overall, the system performed well. Visually, there is a noticeable difference between the experiment with ambient lighting and no ambient lighting. This is further emphasized in the calculations section. In the experiment with no ambient light, the system performed with over two times the precision of the experiment with ambient light. This is shown in the grouping of the scatterplots, as well as in the standard deviation figures. Additionally, the experiment with no ambient light performed with approximately six times the accuracy of the experiment with ambient light. Once again, this is shown both in the scatter plot, and the percentage error figures. In both experiments, a decrease in both precision and accuracy can be observed as the range of the target increased. This effect is especially prevalent in experiment one, as the ambient light appears to decrease effective range. Another trend to notice is the grouping of some measurements not over the target. In certain instances, such as experiment two's position two, the dots on the scatterplot are tightly grouped indicating high precision. However, this group of dots is shifted slightly away from the actual position of the target. This could indicate problems with the program, or the need for further calibration.

## Conclusion

Throughout both experiments, infrared TOF sensors have proved to be a viable alternative to traditional tracking means. This study has revealed not only the viability of these sensors, but also shown strengths and weaknesses of this approach, illuminating possible use cases. It is clearly shown in the experiments that the system functions better with no ambient lighting. This data could reveal an application in nighttime tracking for these sensors. This is especially viable when considering the difficulty with the camera and computer vision approach at night (Lie et al, 2021). Further experimentation with ultrasonic sensors on a similar test setup would be beneficial to make accurate comparisons. One potential limitation of this system shown in this study was the decrease in accuracy with increased distance to the target. Further tests could be conducted in order to quantify what distance from the sensor causes this decrease in accuracy. Additionally, accuracy results from this study could be improved with better software calibration. In many trials of the experiments, reported values were y-shifted away from the target. This y-shift could be measured and compensated for in software.